\pdfoutput=1

\documentclass[11pt]{article}

\usepackage{acl}

\usepackage{times}
\usepackage{latexsym}

\usepackage[T1]{fontenc}

\usepackage[utf8]{inputenc}

\usepackage{microtype}

\usepackage{inconsolata}
\usepackage{amsmath}
\usepackage{txfonts}
\usepackage{boldline}
\usepackage{array}
\usepackage{graphicx}
\usepackage{tabularx}
\usepackage{arydshln}
\usepackage{xspace}
\usepackage{supertabular}
\usepackage{multirow}
\usepackage{booktabs}
\usepackage{makecell}
\usepackage{enumitem}
\newcolumntype{C}[1]{>{\centering\arraybackslash}m{#1}}
\newcolumntype{L}[1]{>{\arraybackslash}m{#1}}
\usepackage{kotex} 
\usepackage[export]{adjustbox}
\usepackage{xcolor}
\usepackage[most]{tcolorbox}
\usepackage{float}
\title{ZeroDL: Zero-shot Distribution Learning for Text Clustering\\via Large Language Models}

\author{Hwiyeol Jo$^{1}$, Hyunwoo Lee$^{2}$, Kang Min Yoo$^{1}$, Taiwoo Park$^{3}$\thanks{\ \ Now at Google} \\ 
  $^1$NAVER Cloud, $^2$NAVER, $^3$NAVER Search US\\
  \texttt{hwiyeolj@gmail.com,\{hanu.lee,kangmin.yoo,taiwoo.park\}@navercorp.com} \\
}

\begin{document}
\maketitle
\begin{abstract}
    The advancements in large language models (LLMs) have brought significant progress in NLP tasks. However, if a task cannot be fully described in prompts, the models could fail to carry out the task. In this paper, we propose a simple yet effective method to contextualize a task toward a LLM. The method utilizes (1) open-ended zero-shot inference from the entire dataset, (2) aggregate the inference results, and (3) finally incorporate the aggregated meta-information for the actual task. We show the effectiveness in text clustering tasks, empowering LLMs to perform text-to-text-based clustering and leading to improvements on several datasets. Furthermore, we explore the generated class labels for clustering, showing how the LLM understands the task through data.
\end{abstract}

\section{Introduction}\label{sec:intro}
    Large language models (LLMs) have demonstrated impressive performances on various downstream tasks~\cite{devlin-etal-2019-bert,radford2019language}. These also exhibit the ability to understand the context of input text, known as in-context learning (ICL)~\cite{brown2020language,openai2023gpt4}. ICL allows leveraging LLMs for specific tasks without further extensive training. However, effective use of ICL hinges on well-designed prompts.
    
    While prompts with few-shot examples demonstrably improve performance, they can easily overfit a model to the examples (\citet{perez2021true,mizrahi2023state}; \textit{inter alia}). This led to a growing interest in zero-shot learning, which reduces the need for intricate few-shot selection. Recent advancements in zero-shot learning involve incorporating more sophisticated use of prompt structures, such as Chain-of-Thought~\cite{wei2022chain}, zero-shot reasoning~\cite{kojima2022large}, and models trained to follow instructions~\cite{ouyang2022training,chung2024scaling}. However, how to design prompts for target tasks remains challenging.

    \begin{figure}[t]\centering
        \includegraphics[width=\columnwidth]{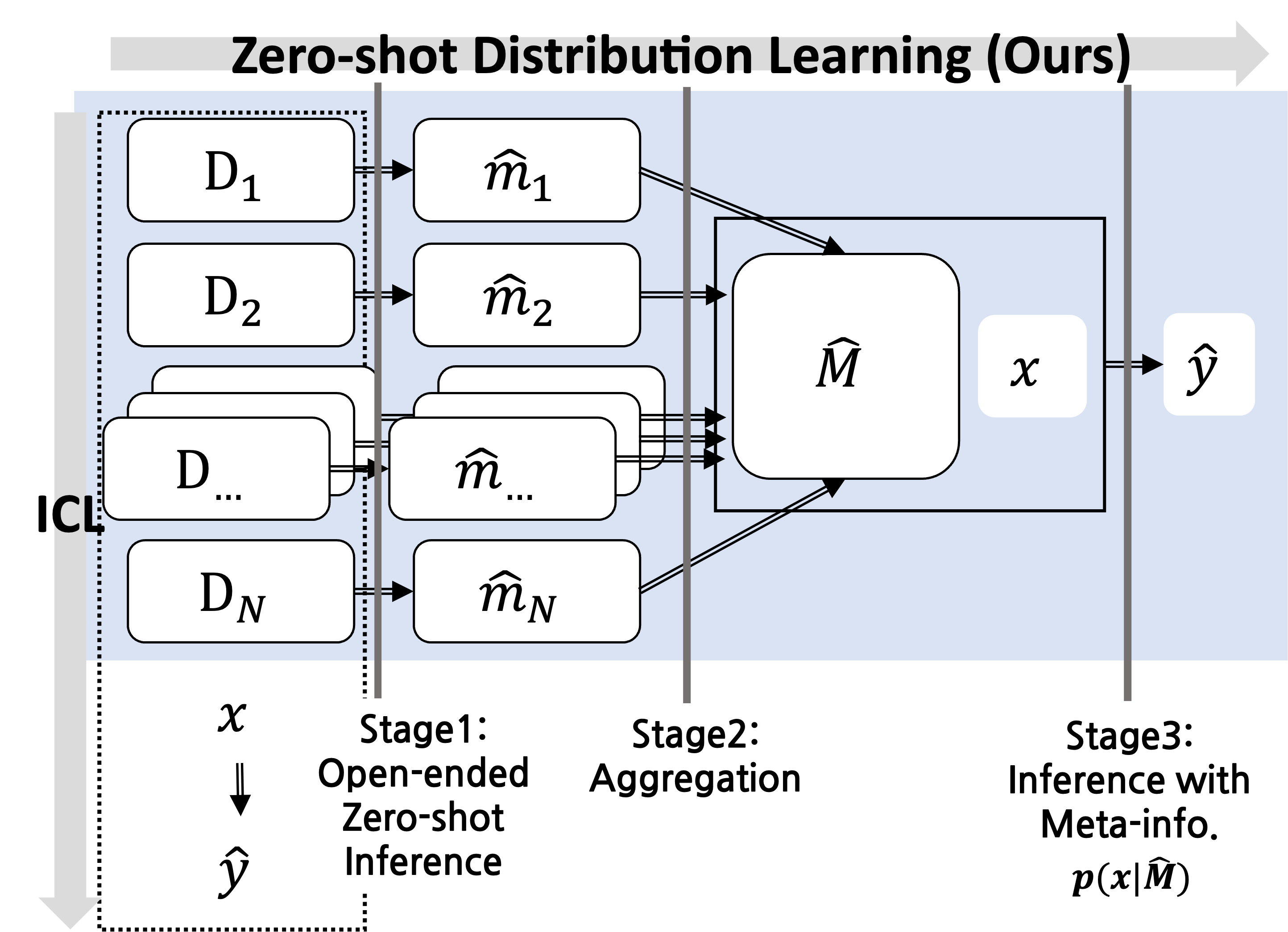}
        \caption{Illustration of proposed method ZeroDL. While in-context learning (ICL) relies on examples (D) tailored to specific tasks, ZeroDL aggregates all the outputs from these zero-shot inferences ($\hat{m}$), resulting in meta-level information ($\hat{M}$). This information is then used by the LLM to generate its final predictions.}\label{fig:framework}
        \vspace{-5mm}
    \end{figure}

    Motivated by the core principle of ICL--providing task and data contexts \textit{within} prompts--we propose an approach to construct more effective zero-shot prompts by understanding how LLMs describe datasets \textit{across} prompting outputs.
    
    As illustrated in Figure~\ref{fig:framework}, {\bf Zero}-shot {\bf D}istribution {\bf L}earning (ZeroDL) aims to learn data distributions through zero-shot inferences. The method comprises two key components: open-ended zero-shot inference and output aggregation. Zero-shot prompts are then constructed with the generated meta information, and used for actual task. This method takes advantage of the self-generated frame of LLMs to successfully carry out a given task. 

    We exemplify the effectiveness of ZeroDL on text clustering tasks where complete task descriptions cannot be provided due to an absence of ground-truth class labels. In addition, our method works in a text-to-text format, allowing clustering with  specific context. For instance, "I love this movie" and "I hate this movie" express opposite sentiment but belong to the same cluster of movie reviews (see Table~\ref{tab:appendix_class_info_free} in the Appendix for potential risk of absence of the perspectives).

    Our contributions in this paper are as follows:

    \begin{itemize}[noitemsep,topsep=1pt]
        \item We propose a novel approach 
        called {\bf Zero}-shot {\bf D}istribution {\bf L}earning 
        that leverages zero-shot inferences to generate meta-level information about the data distribution by aggregating open-ended inference outputs from datasets.
        \item ZeroDL allows models to perform text-based clustering, empowering them to handle data with specific context, which offers advantages over embedding-based clustering methods.
        \item ZeroDL is competitive against embedding-based clustering methods on several datasets. Notably, ZeroDL even achieves better performances than models with ground-truth class labels in some cases.
    \end{itemize}

\section{Related Works}\label{sec:related_work}
    The majority of existing LLM-based clustering approaches rely on traditional methods like K-Means, employing LLM-generated embeddings as input~\cite{petukhova2024text,behnamghader2024llm2vec}. ClusterLLM~\cite{zhang2023clusterllm} introduced LLM-guided refinement of embedding-based clustering models. In contrast, our approach ZeroDL, directly leverages text-level prompting to inject specific viewpoints into LLMs, enabling more targeted and contextualized clustering.

    Other works have explored using LLMs as clustering models~\cite{wang2023goal,viswanathan2023large,pham2023topicgpt,huang2024text}, but these approaches typically relied on few-shot settings with access to gold labels. IDAS~\cite{de2023idas} addressed this limitation by selecting representative data using an auxiliary embedding model to generate pseudo-labels without demonstrations. These labels are subsequently refined through in-context learning.

    ZeroDL distinguishes itself by operating in a fully zero-shot setting, eliminating the need for additional embedding models. Moreover, while IDAS's class labels are primarily influenced by a limited set of data points and generate the labels independently, ZeroDL considers the entire data distribution through an aggregation step. This enables our method to generate class labels in an auto-regressive manner, taking into account the relationships between different labels.
    
    The importance of appropriate ground-truth labels extends beyond clustering and permeates ICL. While \citet{min-etal-2022-rethinking} observed cases where the input-label correspondence does not play significant roles, \citet{yoo-etal-2022-ground} argued that the impact heavily depends on target tasks and experiment settings. We believe that this work would serve as a reference that appropriate class labels can be successfully generated by LLMs themselves.

\section{Proposed Method: ZeroDL}\label{sec:method}

    \paragraph{Stage 1: Open-Ended Inference}
    We begin by designing a prompt for zero-shot classification. This prompt intentionally avoids any detailed information about the task, minimizing the risk of overfitting. Based on the idea, we opt for the simplest prompt format:
    \begin{tcolorbox}[colback=gray!10, colframe=black!75, title=\small Open-Ended Inference Prompt Template]\small
        Text: {\tt [text]}
        \\
        \\
        Classify the text to the
        \small best {\tt [type\_of\_task]} class.
    \end{tcolorbox}
    
    \noindent where {\small\tt [text]} is the input data and {\small\tt [type\_of\_task]} provides view of the task. In the experiment, it can be either sentiment or topic. Leveraging this prompt, we perform model inferences on all the input data ($D$). This process generates open-ended class predictions, which will be denoted as $\hat{m}$.
    
\paragraph{Stage 2: Aggregation}
    The open-ended predictions lack constraints, leading to potentially inconsistent output formats. For instance, the model might predict "positive" and "non-positive" classes, while the ground-truth is "positive" and "negative". The predictions could even be entire sentences.
    To address the inconsistencies in the open-ended predictions, we employ aggregation strategy.
    
    Before the aggregation, we count the frequency of each predictions and sort it, denoted as $\hat{m}_1, \hat{m}_2, \cdots, \hat{m}_U$ where $U$ denotes the unique number of predictions and the frequency of $\hat{m}_{n}$ is equal or larger than $\hat{m}_{n+1}$. After that, the predictions which frequency is only 1 are dropped in order to remove extraordinary predictions
    and save computation. Next, we iteratively construct subsets of the predictions by removing the least frequent predictions one by one. This process results in a list of subsets, denoted as $S$; $S_U = \{\hat{m}_1, \hat{m}_2, \cdots, \hat{m}_U\}$, $S_{U-1} = \{\hat{m}_1, \hat{m}_2, \cdots, \hat{m}_{U-1}\}$, $\cdots$, $S_1 = \{\hat{m}_1\}$. We input each subset to the LLM, providing more weights to the frequently occurred predictions (e.g., $\hat{m}_1$ is repeatedly included in the subsets while $\hat{m}_U$ occurs only once).
    The prompt template for aggregation is as follows:

    \begin{tcolorbox}[colback=gray!10, colframe=black!75, title=\small Aggregation Prompt Template]\small
        {\tt [type\_of\_task]} List:
        \begin{tcbraster}[raster equal height, raster valign=top, raster columns=4, raster column skip=1mm, raster rows=1, colback=gray!10,colframe=black!75
    ]
    \begin{tcolorbox}[boxrule = 0.5pt,colback=white!50,colframe=gray!30,title=\small\tt $S_U$,width=0.5cm,coltitle=gray]
        {\small\tt $\hat{m_1}$\\
        $\hat{m_2}$\\
        $\cdots$\\
        $\hat{m}_{U-1}$\\
        $\hat{m_U}$}
    \end{tcolorbox}
    \begin{tcolorbox}[boxrule = 0.5pt,colback=white!50,colframe=gray!30,title=\small\tt $S_{U-1}$,width=0.5cm,coltitle=gray]
        {\small\tt $\hat{m}_1$\\
        $\hat{m}_2$\\
        $\cdots$\\
        $\hat{m}_{U-1}$}\\
    \end{tcolorbox}
    \begin{tcolorbox}[boxrule = 0.5pt,colback=white!50,colframe=gray!30,title=\small\tt $S_{\cdots}$,width=0.5cm,coltitle=gray]
        {\small\tt $\hat{m}_1$\\
        $\hat{m}_2$\\
        $\cdots$}\\
    \end{tcolorbox}
    \begin{tcolorbox}[boxrule = 0.5pt,colback=white!50,colframe=gray!30,title=\small\tt $S_{1}$,width=0.5cm,coltitle=gray]
        {\small\tt $\hat{m}_1$}
    \end{tcolorbox}
    \end{tcbraster}
    
    Aggregate the {\tt [type\_of\_task]} List into
        {\tt [NUM\_CLUSTER\_CLASS] classes.}
    \end{tcolorbox}
    \noindent where {\small\tt [NUM\_CLUSTER\_CLASS]} is pre-defined number of classes to cluster.
    However, the model outputs often still lack coherence, especially in generating exact number of classes. To address this, we select the generation results only when the number of aggregated classes matches the pre-defined number of cluster classes.
    We then use the most frequent aggregated classes
    as meta-information $\hat{M}$ (class labels in the experiments). This information represents the model's understanding of potential views over the entire data.

\paragraph{\bf Stage 3: Leveraging Meta-Information}
    We incorporate the aggregated meta-information ($\hat{M}$) into the original prompt\footnote{The order of {\small Text:} and {\small Class description:} can be reversed.}:
    \begin{tcolorbox}[colback=gray!10, colframe=black!75, title=\small Final Prediction Prompt Template]\small
        Text: {\tt [text]}\\
        \\
        Class description:\\
        - Class 0: $\hat{M_0}$\\
        - Class $\cdots$: $\hat{M_{\cdots}}$\\
        - Class $C$: $\hat{M_{C}}$\\
        \\
        Based on the class description, classify the text to the best {\tt [type\_of\_task]} class.
    \end{tcolorbox}

    \begin{table*}[t]\centering\small
    \scalebox{0.9}{
        \begin{tabular}{@{}
        l@{\hskip 0.18cm}
        p{2.2cm}@{\hskip 0.18cm}p{0.8cm}@{\hskip 0.18cm}p{0.8cm}@{\hskip 0.18cm}p{0.8cm}@{\hskip 0.18cm}p{0.8cm}@{\hskip 0.18cm}p{0.8cm}@{\hskip 0.18cm}p{0.8cm}@{\hskip 0.18cm}p{0.8cm}@{\hskip 0.18cm}p{0.8cm}@{\hskip 0.18cm}p{0.8cm}@{\hskip 0.5cm}p{1.0cm}@{\hskip 0.18cm}p{1.0cm}}
        \toprule
        {\tt Model} & {\tt Method} & {\tt IMDB} & {\tt SST-2} & {\tt SST-5} & {\tt YRev} & {\tt AGNew} & {\tt DBp(F)} & {\tt DBp(B)} & {\tt Yah(F)} & {\tt Yah(B)} & {\tt Macro} & {\tt Micro} \\
        \midrule
        \midrule
        \tt hcx-seed & \tt ZeroDL(C-T) & 56.2 & 62.5 & \underline{33.6} & \underline{\bf 43.6} & \underline{\bf 33.1} & \underline{24.1} & 16.9 & \bf 21.3 & \bf 18.3 & 34.4 & 33.1 \\
        \tt -0.5b-it & \tt IDAS(C-T) & \underline{88.3} & \underline{74.9} & 27.7 & 35.0 & 29.4 & 17.0 & \underline{26.8} & - & - & - & - \\
        \cmidrule{2-13}
        & \tt Gold(C-T) & \bf 72.1 & \bf 75.2 & \bf 35.2 & 27.6 & 31.4 & 31.2 & 24.1 & 18.3 & \bf 18.3 & \bf 37.0 & \bf 33.6 \\
        \cmidrule{2-13}
        & \tt ZeroDL(T-C) & 86.7 & \underline{72.2} & \underline{\bf 36.0} & \underline{\bf 43.9} & \underline{\bf 59.5} & \underline{32.3} & \underline{44.5} & 26.7 & 22.8 & 47.2 & 46.4 \\
        & \tt IDAS(T-C) & \underline{\bf 89.4} & 62.9 & 28.9 & 28.7 & 27.3 & 23.7 & 27.2 & - & - & - & - \\
        \cmidrule{2-13}
        & \tt Gold(T-C) & 89.1 & \bf 73.7 & 34.4 & 39.1 & 59.4 & \bf 61.1 & \bf 57.2 & \bf 40.3 & \bf 56.2 & \bf 56.7 & \bf 56.3 \\
        \midrule
        \tt mistral & \tt llm2vec+KMeans & 62.1 & 55.2 & 30.3 & \underline{\bf 56.0} & \underline{\bf 84.4} & \underline{\bf 96.1} & \underline{70.8} & \underline{\bf 48.0} & 46.2 & 61.0 & \underline{67.5} \\
        \cmidrule{2-13}
        \tt -7b-it & \tt ZeroDL(C-T) & \underline{\bf 88.8} & \underline{\bf 85.7} & 40.5 & 48.7 & 75.2 & 58.9 & 64.1 & 47.2 & \underline{61.4} & \underline{63.4} & 61.5 \\
        & \tt IDAS(C-T) & 51.1 & 63.1 & \underline{41.4} & 37.3 & 38.1 & 32.1 & 21.8 & 30.2 & 41.8 & 39.7 & 35.4 \\
        \cmidrule{2-13}
        & \tt Gold(C-T) & 87.5 & 77.0 & \bf 41.8 & 50.8 & 60.7 & 74.2 & {\bf 85.2} & 40.8 & \bf 62.5 & \bf 64.5 & \bf 68.1 \\
        \cmidrule{2-13}
        & \tt ZeroDL(T-C) & \underline{90.2} & \underline{84.2} & 36.0 & 46.8 & 79.5 & 56.7 & 72.2 & \underline{\bf 51.0} & \underline{66.3} & \underline{64.8} & 63.0 \\
        & \tt IDAS(T-C) & 50.1 & 68.1 & \underline{39.6} & 41.0 & 44.6 & 35.8 & 27.6 & 33.3 & 46.1 & 42.9 & 38.9 \\
        \cmidrule{2-13}
        & \tt Gold(T-C) & \bf 91.7 & {\bf 82.5} & \bf 43.3 & 51.8 & 82.7 & 84.1 & {\bf 82.7} & 50.9 & {\bf 73.8} & {\bf 71.5} & {\bf 72.7} \\
        \midrule
        \tt Qwen-2.5 & \tt ZeroDL(C-T) & 87.6 & 74.8 & \underline{\bf 45.5} & \underline{\bf 53.8} & \underline{\bf 70.9} & \underline{62.3} & \underline{75.9} & \underline{45.3} & \underline{\bf 68.8} & \underline{65.0} & \underline{65.8} \\
        \tt -7b-it & \tt IDAS(C-T) & \underline{\bf 93.3} & \underline{\bf 87.3} & 42.7 & 51.1 & 68.8 & 47.4 & 68.4 & 30.3 & 41.8 & 58.9 & 59.0 \\
        \cmidrule{2-13}
        & \tt Gold(C-T) & 84.3 & \bf 87.4 & 45.0 & 49.9 & 62.2 & \bf 84.5 & \bf 89.0 & \bf 51.9 & 68.2 & \bf 69.2 & \bf 71.3 \\
        \cmidrule{2-13}
        & \tt ZeroDL(T-C) & 93.5 & 85.2 & \underline{\bf 50.5} & 49.3 & \underline{\bf 84.3} & \underline{66.7} & \underline{78.3} & \underline{50.8} & \underline{74.4} & \underline{70.3} & \underline{68.2} \\
        & \tt IDAS(T-C) & \underline{\bf 95.3} & \underline{\bf 88.9} & 45.1 & \underline{\bf 51.6} & 83.1 & 56.0 & 72.2 & 31.6 & 43.3 & 63.0 & 62.7 \\
        \cmidrule{2-13}
        & \tt Gold(T-C) & 94.7 & 88.5 & 46.3 & 49.0 & 82.9 & \bf 94.2 & \bf 98.0 & \bf 64.5 & \bf 79.7 & \bf 77.5 & \bf 78.6 \\
        \midrule
        \tt gemma-2 & \tt ZeroDL(C-T) & 94.3 & 88.0 & \underline{\bf 52.6} & 51.6 & \underline{\bf 74.5} & 52.8 & \underline{53.5} & 50.3 & 60.0 & 64.2 & 60.1 \\
        \tt -27b-it & \tt IDAS(C-T) & \underline{\bf 95.3} & \underline{\bf 91.1} & 42.7 & \underline{\bf 58.3} & 31.7 & \underline{56.9} & 44.0 & - & - & - & - \\
        \cmidrule{2-13}
        & \tt Gold(C-T) & 79.8 & 86.3 & 48.7 & 57.3 & 70.5 & {\bf 89.8} & {\bf 97.1} & {\bf 56.1} & {\bf 65.9} & \bf 72.4 & \bf 75.9 \\
        \cmidrule{2-13}
        & \tt ZeroDL(T-C) & 94.0 & 88.3 & \underline{50.8} & 50.8 & \underline{81.8} & 57.3 & \underline{57.6} & 52.7 & 73.0 & 67.4 & 62.7 \\
        & \tt IDAS(T-C) & \underline{\bf 95.2} & \underline{89.2} & 42.4 & \underline{\bf 55.6} & 35.0 & \underline{60.5} & 49.8 & - & - & - & - \\
        \cmidrule{2-13}
        & \tt Gold(T-C) & 95.3 & {\bf 89.6} & {\bf 51.8} & {\bf 55.7} & {\bf 85.4} & {\bf 92.6} & {\bf 98.7} & {\bf 66.9} & {\bf 84.4} & \bf 80.0 & \bf 81.0 \\
        \bottomrule
        \end{tabular}
    }
    \caption{\label{tab:main_result}
    The performance of ZeroDL for text clustering. {\small\tt C-T} denotes the prompt order with class information then input text. {\small\tt T-C} is the reversed. Bold means the best accuracy in the same prompt order and underline denotes the outperforming cases than baselines (except for Gold label). IDAS with HyperCLOVA X SEED and Gemma-2 is constrained by the maximum 8,192 token length. More baselines are presented in Table~\ref{tab:main_result_appendix} of the Appendix.}
    \vspace{-4mm}
    \end{table*}

    \begin{table*}[t] \centering\small
    \scalebox{0.9}{
        \begin{tabular}{c||c|c}
        \hlineB{3}
            {\tt Data} & \tt Method & \tt ClassLabels \\
        \hlineB{3}
        
        \multirow{3}{*}{{\tt SST-5}} & \tt ZeroDL &
            \begin{tabularx}{0.9\linewidth}{X}
                {{\bf Neutral Sentiment}{\color{gray}: This class includes all the sentiment labels that express a neutral sentiment towards the movie or documentary. Examples include (...) }} \\ 
                {{\bf Negative Sentiment}{\color{gray}: This class includes all the sentiment labels that express a negative or sad emotion towards the movie or documentary. Examples include (...) }} \\ 
                {{\bf Ambiguous Sentiment}{\color{gray}: This class includes all the sentiment labels that do not clearly express a positive or negative emotion towards the movie or documentary. Examples include (...) }} \\ 
                {{\bf Mixed Sentiment}{\color{gray}: This class includes all the sentiment labels that express a mixed sentiment towards the movie or documentary. Examples include (...) }} \\ 
                {{\bf Positive Sentiment}{\color{gray}: This class includes all the sentiment labels that express a positive emotion towards the movie or documentary. Examples include (...) }} \\ 
            \end{tabularx} \\
            \cline{2-3}
                & {\tt IDAS} &
            \begin{tabularx}{0.90\linewidth}{X}
                {{\bf Disappointing}, {\bf Critical}, {\bf Negative}, {\bf Positive}, {\bf Disappointing or Unsatisfying}}
            \end{tabularx} \\
            \cline{2-3}
                                & {\tt Gold} &
            \begin{tabularx}{0.90\linewidth}{X}
                {{\bf Very Negative}, {\bf Negative}, {\bf Neutral}}, 
                {{\bf Positive}, {\bf Very Positive}} \\
            \end{tabularx} \\
            \cline{2-3}
        \hline
        \multirow{5}{*}{{\tt AGNews}}    & \tt ZeroDL &
            \begin{tabularx}{0.90\linewidth}{X}
                {{\bf International Relations and Politics}{\color{gray}: This class includes topics related to international relations, diplomacy, Middle East politics, terrorism, nuclear politics, and elections.}} \\
                {{\bf Sports and Entertainment}{\color{gray}: This class includes topics related to sports, tennis, golf, basketball, baseball, cricket, and entertainment.}} \\
                {{\bf Business and Economy}{\color{gray}: This class includes topics related to the economy, finance, stocks, mergers and acquisitions, retail, real estate, and labor markets.}} \\
                {{\bf Technology and Science}{\color{gray}: This class includes topics related to technology, computing, internet, cybersecurity, space exploration, and science.}} \\
                \end{tabularx} \\
            \cline{2-3}
            & {\tt IDAS} &
            \begin{tabularx}{0.90\linewidth}{X}
                {\bf Iran's Nuclear Program, Oil Prices, Businesses prioritize hardware upgrades in economic recovery, Volcanic activity at Mount St. Helens}
            \end{tabularx} \\
            \cline{2-3}
                                & {\tt Gold} &
            \begin{tabularx}{0.90\linewidth}{X}
                {{\bf World}, {\bf Sports}, {\bf Business}, {\bf Sci/Tech}}
            \end{tabularx} \\
        \hlineB{3}
        \end{tabular}
    }
    \caption{The example of generated class labels in 
    5-class sentiment classification ({\small\tt SST-5}), and topic classification ({\small\tt AGNews}). ZeroDL can generates alternative class labels and its description. Compared with IDAS that has overlap in the generated labels, ZeroDL considers other labels in generation. Furthermore, IDAS's representative data points fail to generate general topic class labels.
    Additional examples are in Table~\ref{tab:appendix_additional_class_gen}.}
    \label{tab:class_info_analysis}
    \vspace{-3mm}
    \end{table*}
    

    By incorporating the meta-information ($\hat{M}$) into the prompt, we enable the LLM to perform conditioned classification within the clustering context.

\section{Experiments}
    \paragraph{Models}
    Our primary experiments utilize {\small\tt mistral-7b-instruct-v0.2}~\cite{jiang2023mistral}. We further evaluate our approach with several other models, including Qwen-2.5~\cite{qwen2.5}, Gemma-2~\cite{team2024gemma}, Llama-3.1~\cite{llama3}. Lastly, HyperCLOVA X SEED~\cite{yoo2024hyperclova} demonstrates the result of a relatively small, instruct-following language model. All the results are averaged over 5 runs.

    \paragraph{Baselines}
    IDAS~\cite{de2023idas} represents a zero-shot clustering approach. We employ 8 demonstrations as described in the paper, using SBERT~\cite{opitz-frank-2022-sbert} as an additional embedder. Final prediction is performed using the same prompt template as our Stage 3.
    We also include llm2vec~\cite{behnamghader2024llm2vec} with K-Means where applicable. This method represents a baseline using (not exactly but) the same backbone computed by embedding-level\footnote{Note that IDAS requires an additional embedding model and llm2vec involves additional training steps.}. We exclude other few-shot settings from our comparisons as they are not considered fair due to the inherent advantages of leveraging additional information from demonstrations and the subjectivity involved in the few-shot selection.
    
    \paragraph{Setting}
    Our method aims to effectively cluster data distributions, converging them into a suitable number of classes. We recognize that clustering datasets often exhibit a high number of classes relative to the available data points. To address this, we utilize text classification datasets, which typically offer a larger number of data instances per class than clustering datasets. Moreover, the availability of ground-truth labels facilitates direct comparison between generated and actual labels, enabling qualitative analysis and interpretability.

    \paragraph{Datasets}
    We thus use 6 text classification datasets for clustering: IMDB~\cite{maas-EtAl:2011:ACL-HLT2011}, SST-2, SST-5~\cite{socher2013recursive}, YelpReivews~\cite{zhang2015character} for sentiment classification, and AGNews, DBpedia~\cite{lehmann2015dbpedia}, YahooAnswers~\cite{chang2008importance} for topic classification.
    Data information is presented in Table~\ref{tab:appendix_data_info} of the Appendix.
    
    \paragraph{Evaluation}
    Model performance is evaluated based on the highest accuracy achieved across all possible mapping combinations. To determine the predicted class, we leverage the {\small\tt Class n} anchor tokens within the LLM outputs.
    However, LLMs might not directly predict the same classes as the ground-truth labels\footnote{For example, {\small\tt Class 0} means {\small\tt Positive} in the prediction while {\small\tt Negative} is labeled as {\small\tt Class 0} in the ground-truth.}, so we test all possible mapping combinations and report the best performed one. We split datasets with more than 7 classes (i.e., DBpedia and YahooAnswers) into 2 subsets Front (F) and Back (B) to avoid out-of-memory\footnote{The computation is a factorial of the number of classes.} after removing the 3 smallest size of classes\footnote{CarsAndTransportation, SocialScience, and Sports classes in YahooAnswers dataset.}.
    
\section{Results}
    Table~\ref{tab:main_result} presents the performance of our ZeroDL method.
    Notably, ZeroDL surpasses models provided with ground-truth (Gold) class labels on several datasets. This suggests that ZeroDL may uncover richer or more nuanced class labels within the data compared with pre-defined labels. Our method demonstrates performance comparable to K-Means clustering using LLM embeddings, particularly excelling on datasets with relatively smaller sizes.
    ZeroDL outperforms IDAS, especially on datasets with more than 2 classes. IDAS often struggles to generate appropriate class labels, particularly when relying on a limited coverage of representative data instances as demonstrations. Furthermore, IDAS is susceptible to the input length limitations of LLMs when the demonstrations are long, as evident in the Gemma results. In contrast, ZeroDL achieves its results through flexible zero-shot prompting without requiring any modifications.

    Table~\ref{tab:class_info_analysis} shows examples of class labels generated from ZeroDL. These labels often provide richer and more informative explanations\footnote{Note that the generation of class description varies depending on datasets and LLMs.} compared with the original ground-truth labels. Moreover, ZeroDL can potentially uncover novel classes based on the data, such as Ambiguous Sentiment and Mixed Sentiment\footnote{Examples are provided in  Table~\ref{tab:appendix_new_class_example} of the Appendix.}. This highlights the limitations of pre-defined labels in capturing the full spectrum of sentiment complexity.
    
    \begin{figure}[t]\centering
        \includegraphics[scale=0.3]{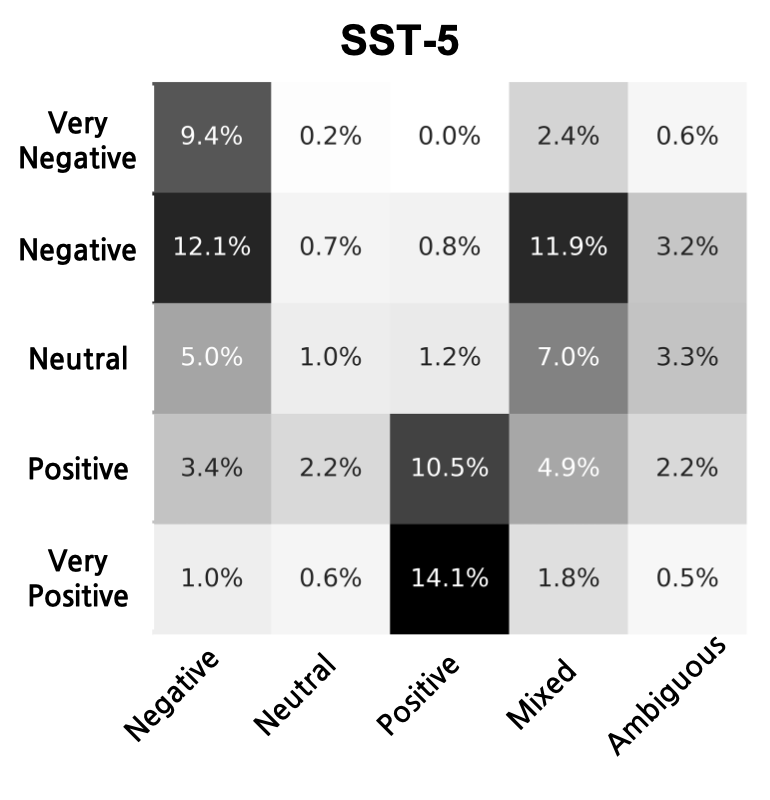}
        \hspace{-3mm}
        \includegraphics[scale=0.29]{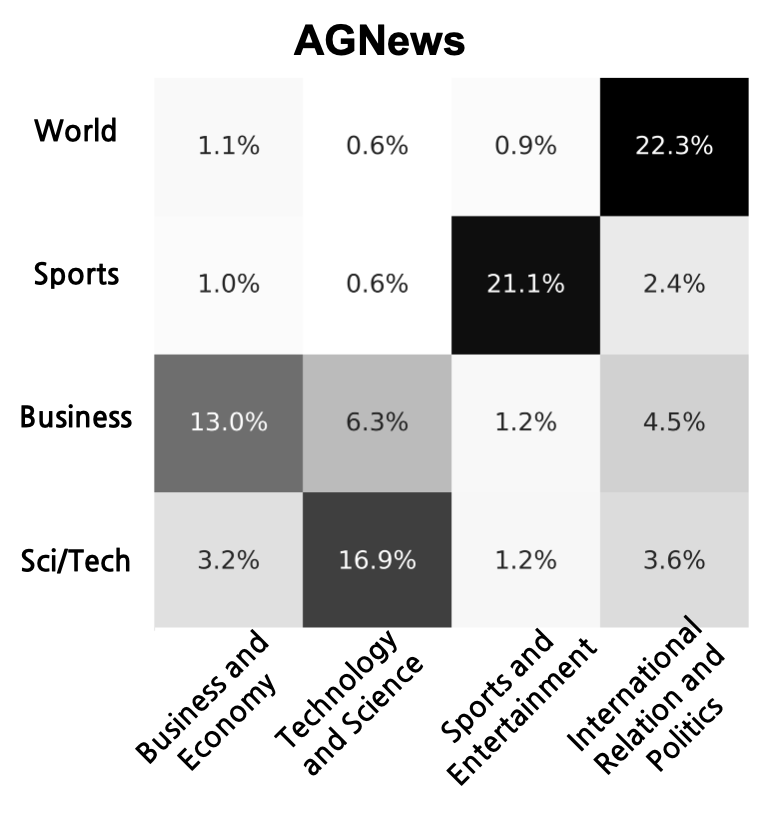}
        \vspace{-4mm}
        \caption{Confusion matrix of proposed method ZeroDL in SST-5 (left) and AGNews (right). x-axis and y-axis are generated labels and gold labels, respectively.}\label{fig:analysis}
        \vspace{-3mm}
    \end{figure}
    
    Figure~\ref{fig:analysis} illustrates the confusion matrix of the results. The relatively low performance of SST-5, even with gold labels, underscores the inherent difficulty of sentiment classification. In contrast, AGNews demonstrates successful matching with our method. This observation emphasizes the potential value of ZeroDL in scenarios where ground-truth labels are unavailable.

    Table~\ref{tab:auto_label_comparison} investigates the significance of class labels in text clustering tasks. We explore the performance of {\small\tt mistral-7b-instruct-v0.2} using class labels suggested by AutoL~\cite{gao-etal-2021-learning} designed for prompt-based model fine-tuning. These labels represent a curated selection.
    We also investigate the class labels generated by {\small\tt gpt-4.1-mini}.
    The results demonstrate that clustering performance is generally higher when using manually selected labels than using the original dataset labels. This suggests that carefully chosen labels can significantly improve clustering outcomes. In the context, ZeroDL performs particularly well on SST-5, implying that the potential of ZeroDL to capitalize on informative class labels automatically.
    
    ZeroDL involves a trade-off in computational cost. We evaluate the impact of varying input data amounts on performance (see Table~\ref{tab:appendix_data_sampling} in the Appendix). While using only 10\% of the data yields reasonable performance, it can lead to inconsistencies in model performance as evidenced by increased standard deviation.
    

    \begin{table}[t]\centering\small
    \scalebox{0.90}{
    \begin{tabular}{@{}
    l@{\hskip 0.1cm}
    p{1.2cm}@{\hskip 0.18cm}p{1.0cm}@{\hskip 0.18cm}p{1.0cm}@{\hskip 0.18cm}p{1.0cm}@{\hskip 0.18cm}p{1.0cm}}
    \toprule
    & {\tt S2(C-T)} & {\tt S2(T-C)} & {\tt S5(C-T)} & {\tt S5(T-C)} \\
    \midrule
    {\tt RandToken} & 51.6 & 59.5 & 28.6 & 28.6 \\
    \tt AutoL(Best) & {\bf 86.8} & {\bf 84.9} & \bf 43.9 & 40.5 \\
    \tt AutoL(Worst) & 82.5 & 80.7 & 41.1 & 39.7 \\
    \hdashline[0.4pt/2pt]
    {\tt Gold} & 77.0 & 82.5 & 41.8 & {\bf 43.3} \\
    \midrule
    \tt ZeroDL(Mistral) & 80.6 & 83.7 & 40.3 & 40.5 \\
    \tt ZeroDL(gpt-4.1) & 77.0 & 82.5 & 41.0 & 40.3 \\
    
    \bottomrule
    \end{tabular}
    }
    \caption{\label{tab:auto_label_comparison}
    The mistral-7b performance with various class labels. {\small\tt ZeroDL(Mistral)} uses only the generated class titles for fair comparison. The best labels from {\small\tt AutoL} in SST-2 are {\small [Wonderful, Bad]} and the worst are {\small [Irresistible, Pathetic]}. In SST-5, {\small [Terrible, Better, Good, Extraordinary, Unforgettable]} are the best while {\small [Awful, Better, Hilarious, Perfect, Wonderful]} are the worst. {\small\tt gpt-4.1-mini} generates {\small [Positive, Negative]} (the same as Gold) and {\small [Positive, Negative, Neutral, Mixed, Other / Unclear]}, respectively.}
    \vspace{-3mm}
    \end{table}
    
\section{Conclusion}

    We introduce ZeroDL, a novel approach 
    to contextualize tasks for a given LLM. 
    ZeroDL employs open-ended zero-shot
    inference and output aggregation to learn data distributions. We demonstrate its effectiveness, showing competitive performances against embedding-based clustering methods and superior performance than ground-truth labels in some cases. Beyond its clustering capabilities, ZeroDL offers the generation of informative class labels that provide deeper insights into LLMs.
    
\clearpage
\section{Limitations}\label{sec:Limitations}

    \paragraph{Prompt Dependency and Heuristics} ZeroDL relies on carefully designed prompts to guide LLMs towards effective clustering. While our focus was on using simple and intuitive prompts, prompt selection can potentially influence the model's behavior and introduce biases. Future work could explore more sophisticated prompt engineering techniques to further enhance ZeroDL's performance.

    \paragraph{Experiments with Diverse LLMs and Prompts} While we acknowledge the computational limitations (and price) of ZeroDL, investigating its behavior with a wider range of LLMs (including commercial models like GPT, Claude, and Gemini) and prompt templates could provide valuable insights into the generalizability and robustness of the approach.
    


    \paragraph{Lower than State-of-the-Art Performance}
    Achieving state-of-the-art performance is not the sole focus of ZeroDL but it offers a valuable framework in understanding data distributions with zero-shot inference via LLMs. By addressing the limitations mentioned above, ZeroDL has the potential to become a powerful and versatile tool not only for text clustering but data exploration.

    \paragraph{Expensive Computational Cost in Inferences.} Although ZeroDL have an alternative approach to reduce computational burden through data sampling, effectively sampling data to generate appropriate class labels remains a challenge. Although we report the trade-off in computational cost and performance in Table~\ref{tab:appendix_data_sampling} of the Appendix, the technique within the framework presents a valuable future direction.

\section*{Acknowledgement}    
    The authors would like to thank reviewers in several rounds of discussion.
    Thanks to Alice Lee for her help with English writing and Minji Hong for her support in writing this work. Lastly, we extend our gratitude to
    Hyperscale AI Team at NAVER Cloud, for unwavering support in connecting us with key resources throughout this research.
    
\bibliography{anthology,custom}
\bibliographystyle{acl_natbib}

\clearpage
\onecolumn
\appendix

\section{Appendix}
\subsection{Data Information}
    \begin{table*}[!hbt]\centering\small
    \scalebox{1.}{
        \begin{tabular}{@{}
        l@{\hskip 0.2cm}
        p{1.0cm}@{\hskip 0.18cm}p{1.0cm}@{\hskip 0.18cm}p{1.0cm}@{\hskip 0.18cm}p{1.0cm}@{\hskip 0.18cm}p{8cm}}
        \toprule
        & \tt \#Train & \tt \#Valid & \tt \#Test & \tt \#Class & \tt Pre-defined ClassTitle \\
        \midrule
        \tt IMDB & 25,000 & 3,750 & 25,000 & 2 & Negative, Positive \\
        \midrule
        \tt SST-2 & 9,645 & 1,101 & 2,210 & 2 & Negative, Positive \\
        \midrule
        \tt SST-5 & 9,645 & 1,101 & 2,210 & 5 & Very Negative, Negative, Neutral, Positive, Very Positive \\
        \midrule
        \tt YelpReviews & 650,000 & 97,500 & 49,999 & 5 & Very Negative, Negative, Neutral, Positive, Very Positive \\
        \midrule
        \tt AGNews & 120,000 & 18,000 & 7,600 & 4 & World, Sports, Business, Sci/Tech \\
        \midrule
        \tt DBpedia(F) & 280,000 & 41,939 & 35,000 & 7 & Company, EducationalInstitution, Artist, Athlete, OfficeHolder, MeanOfTransportation, Building \\
        \midrule
        \tt DBpedia(B) & 280,000 & 42,213 & 35,000 & 7 & NaturalPlace, Village, Animal, Plant, Album, Film, WrittenWork \\
        \midrule
        \tt Yahoo(F) & 59,518 & 8,879 & 10,489 & 7 & ArtsAndHumanities, BeautyAndStyle, BusinessAndFinance, ComputersAndInternet, ConsumerElectronics, EducationAndReference, EntertainmentAndMusic \\
        \midrule
        \tt Yahoo(B) & 59,493 & 8,973 & 10,514 & 7 & FoodAndDrink, GamesAndRecreation, Health, HomeAndGarden, Pets, PregnancyAndParenting, SocietyAndCulture \\
        \bottomrule
        \end{tabular}
    }
    \caption{\label{tab:appendix_data_info}
    Data information used in the experiments.
    }
    \end{table*}

\subsection{Additional Model Performances}

    \begin{table*}[!hbt]\centering\small
    \scalebox{1.}{
        \begin{tabular}{@{}
        l@{\hskip 0.18cm}
        p{2.2cm}@{\hskip 0.18cm}p{0.8cm}@{\hskip 0.18cm}p{0.8cm}@{\hskip 0.18cm}p{0.8cm}@{\hskip 0.18cm}p{0.8cm}@{\hskip 0.18cm}p{0.8cm}@{\hskip 0.18cm}p{0.8cm}@{\hskip 0.18cm}p{0.8cm}@{\hskip 0.18cm}p{0.8cm}@{\hskip 0.18cm}p{0.8cm}@{\hskip 0.5cm}p{1.0cm}@{\hskip 0.18cm}p{1.0cm}}
        \toprule
        {\tt Model} & {\tt Method} & {\tt IMDB} & {\tt SST-2} & {\tt SST-5} & {\tt YRev} & {\tt AGNew} & {\tt DBp(F)} & {\tt DBp(B)} & {\tt Yah(F)} & {\tt Yah(B)} & {\tt Macro} & {\tt Micro} \\
        \midrule
        \tt TF-IDF & \tt KMeans & 52.1 & 53.0 & 26.2 & 35.2 & 43.8 & 55.9 & 62.8 & 44.7 & 46.5 & 46.7 & 48.8 \\
        \midrule
        \tt SBERT &  \tt KMeans & 65.0 & 54.0 & 33.9 & 30.3 & 82.7 & 95.4 & 93.9 & 67.2 & 68.3 & 65.6 & 67.5 \\
        \midrule
        \tt llama-3.1 & \tt ZeroDL(C-T) & 76.0 & \bf 73.8 & 35.0 & \underline{\bf 51.1} & 54.5 & 37.7 & 58.5 & \underline{46.4} & \underline{47.9} & \underline{53.4} & 53.2\\
        \tt -8b-it & \tt IDAS(C-T) & \underline{\bf 92.2} & \bf 73.8 & \underline{42.4} & 45.4 & \underline{55.4} & \underline{40.8} & \underline{64.1} & 29.2 & 35.3 & 53.2 & \underline{53.9} \\
        \cmidrule{2-13}
        & \tt Gold(C-T) & 80.4 & 73.5 & \bf 44.6 & 48.7 & \bf 64.1 & \bf 74.4 & \bf 80.8 & \bf 49.6 & \bf 69.5 & \bf 65.1 & \bf 66.7 \\
        \cmidrule{2-13}
        & \tt ZeroDL(T-C) & 94.4 & 79.4 & \underline{39.1} & 49.4 & \underline{77.6} & \underline{46.1} & \underline{68.1} & \underline{53.7} & \underline{60.0} & \underline{63.1} & \underline{61.1} \\
        & \tt IDAS(T-C) & \underline{\bf 94.6} & \underline{\bf 84.5} & 38.4 & \underline{\bf 50.1} & 67.7 & 37.6 & 66.4 & 32.0 & 36.2 & 56.4 & 56.2 \\
        \cmidrule{2-13}
        & \tt Gold(T-C) & 94.4 & 81.5 & \bf 43.4 & \bf 50.1 & \bf 78.9 & \bf 80.8 & \bf 75.1 & \bf 62.8 & \bf 78.3 & \bf 71.7 & \bf 71.2 \\
        \midrule
        \tt Qwen-2.5 & \tt ZeroDL(C-T) & 94.5 & \underline{\bf 90.2} & 44.5 & \underline{\bf 62.6} & \underline{71.3} & 60.8 & \underline{64.4} & \underline{53.4} & \underline{71.3} & \underline{68.1} & \underline{67.5} \\
        \tt -72b-it & \tt IDAS(C-T) & \underline{\bf 95.3} & 88.8 & \underline{47.1} & 50.5 & 68.1 & \underline{62.3} & 57.2 & 35.7 & 55.8 & 62.3 & 61.1 \\
        \cmidrule{2-13}
        & \tt Gold(C-T) & 95.1 & 89.4 & {\bf 51.0} & 57.7 & \bf 81.3 & \bf 94.2 & \bf 97.9 & \bf 69.0 & \bf 84.8 & \bf 80.0 & \bf 81.6 \\
        \cmidrule{2-13}
         & \tt ZeroDL(T-C) & 95.3 & \underline{\bf 90.1} & 45.4 & \underline{\bf 63.3} & \underline{76.6} & \underline{65.2} & \underline{65.4} & \underline{55.1} & \underline{73.6} & \underline{70.0} & \underline{69.4} \\
        & \tt IDAS(T-C) & \underline{\bf 95.9} & 89.9 & \underline{46.3} & 51.2 & 76.4 & 63.2 & 58.9 & 34.5 & 47.7 & 62.7 & 61.7 \\
        \cmidrule{2-13}
        & \tt Gold(T-C) & {\bf 95.8} & {\bf 90.0} & {\bf 52.7} & 59.9 & \bf 83.9 & \bf 96.8 & \bf 99.7 & \bf 73.7 & \bf 87.4 & \bf 82.2 & \bf 83.8 \\ 
        \bottomrule
        \end{tabular}
    }
    \caption{\label{tab:main_result_appendix}
    The performance of ZeroDL for text clustering compared with more methods. {\small\tt C-T} denotes the prompt order with class information then input text. {\small\tt T-C} is the reversed. Bold means the best accuracy in the same prompt order and underline the outperforming cases than baseline except for ground-truth class labels.
    }
    \end{table*}

\clearpage
\subsection{Qualitative Examples}
    \begin{table*}[!hbt]\centering\small
    \scalebox{1}{
        \begin{tabular}{ccl}
        \toprule
        {\tt Predicted} & {\tt Gold} & {\tt Example} \\
        \midrule
        \multirow{6}{*}{\makecell[c]{\tt Mixed\\ \tt Sentiment}} & \tt VeryNeg & \makecell[l]{It 's hard not to feel you 've just watched a feature-length video game\\with some really heavy back story .} \\
        \cline{2-3}
         & \tt Neg & But it pays a price for its intricate intellectual gamesmanship . \\
        \cline{2-3}
         & \tt Neutral & \makecell[l]{The appearance of Treebeard and Gollum 's expanded role will either\\have you loving what you 're seeing, or rolling your eyes .} \\
        \cline{2-3}
         & \tt Pos & \makecell[l]{An utterly compelling ` who wrote it ' in which the reputation of\\the most famous author who ever lived comes into question .} \\
        \cline{2-3}
         & \tt VeryPos & ... a roller-coaster ride of a movie \\
        \midrule
         \multirow{7}{*}{\makecell[c]{\tt Ambiguous\\ \tt Sentiment}} & \tt VeryNeg & \makecell[l]{It 's difficult to say whether The Tuxedo is more boring or embarrassing \\-- I 'm prepared to call it a draw .}\\
        \cline{2-3}
         & \tt Neg & \makecell[l]{Like most Bond outings in recent years ,\\some of the stunts are so outlandish that they border on being cartoonlike .} \\
        \cline{2-3}
         & \tt Neutral & Effective but too-tepid biopic \\
        \cline{2-3}
         & \tt Pos & But he somehow pulls it off . \\
        \cline{2-3}
         & \tt VeryPos & \makecell[l]{Emerges as something rare , an issue movie that 's so honest\\and keenly observed that it does n't feel like one .}\\
        \bottomrule
        \end{tabular}
    }
    \caption{\label{tab:appendix_new_class_example}
    Example of data predicted to newly generated classes in SST-5. Randomly selected.
    }
    \end{table*}

\subsection{Data Sampling Ablation}
    \begin{table*}[!hbt]\centering\small
    \scalebox{1}{
        \begin{tabular}{@{}
        l@{\hskip 0.5cm}
        p{1cm}@{\hskip 0.18cm}p{1cm}@{\hskip 0.18cm}p{1cm}@{\hskip 0.18cm}p{1cm}@{\hskip 0.18cm}p{1cm}@{\hskip 0.18cm}p{1cm}@{\hskip 0.18cm}p{1cm}@{\hskip 0.18cm}p{1cm}@{\hskip 0.18cm}p{1cm}@{\hskip 0.18cm}p{1cm}@{\hskip 0.18cm}p{1cm}}
        \toprule
        {\tt [C-T]} & {\tt IMDB} & {\tt SST-2} & {\tt SST-5} & {\tt YRev} & {\tt AGNew} & {\tt DBp(F)} & {\tt DBp(L)} & {\tt Yah(F)} & {\tt Yah(L)} \\ 
        \midrule
        \tt All & 88.8 & 85.7 & 40.5 & 48.7 & 75.2 & 58.9 & 64.1 & 47.2 & 61.4 \\
        \tt (std) & 3.20 & 0.18 & 1.22 & 0.21 & 1.22 & 7.85 & 6.61 & 1.02 & 1.92 \\
        \midrule
        \tt 1\% & 81.3 & 79.3 & 46.9 & 51.5 & 66.0 & 59.4 & 58.4 & 36.1 & 43.7 \\
        \tt (std) & 2.97 & 1.5 & 0.88 & 0.12 & 5.71 & 1.93 & 5.11 & 0.72 & 3.35 \\
        \tt 5\% & 74.6 & 83.6 & 45.8 & 49.4 & 74.2 & 58.8 & 66.4 & 39.3 & 58.5 \\
        \tt (std) & 18.49 & 1.37 & 2.78 & 0.47 & 1.94 & 5.38 & 9.13 & 4.82 & 2.73 \\
        \tt 10\% & 87.0 & 84.5 & 44.8 & 49.2 & 74.5 & 67.2 & 63.9 & 40.4 & 60.4 \\
        \tt (std) & 5.90 & 1.50 & 5.02 & 0.29 & 1.44 & 6.35 & 8.95 & 5.11 & 2.96 \\
        \bottomrule
        \toprule
        {\tt [T-C]} & {\tt IMDB} & {\tt SST-2} & {\tt SST-5} & {\tt YRev} & {\tt AGNew} & {\tt DBp(F)} & {\tt DBp(L)} & {\tt Yah(F)} & {\tt Yah(L)} \\
        \midrule
        \tt All & 90.2 & 84.2 & 36.0 & 46.8 & 79.5 & 56.7 & 72.2 & 51.0 & 66.3 \\
        \tt (std) & 3.52 & 0.24 & 3.84 & 0.47 & 1.73 & 9.73 & 6.86 & 0.75 & 2.80 \\
        \midrule
        \tt 1\% & 88.4 & 58.6 & 42.0 & 49.8 & 74.6 & 57.7 & 67.5 & 37.7 & 49.4 \\
        \tt (std) & 0.51 & 11.93 & 3.99 & 2.05 & 4.34 & 1.31 & 8.37 & 4.1 & 3.15 \\
        \tt 5\% & 85.5 & 80.2 & 41.7 & 46.2 & 80.7 & 57.1 & 74.1 & 37.2 & 68.1 \\
        \tt (std) & 7.36 & 1.92 & 4.09 & 1.42 & 2.83 & 10.58 & 9.36 & 6.87 & 4.58 \\
        \tt 10\% & 91.2 & 82.0 & 39.1 & 46.6 & 79.9 & 63.2 & 75.3 & 45.2 & 68.5 \\
        \tt (std) & 2.16 & 1.95 & 2.26 & 0.33 & 2.74 & 6.53 & 9.91 & 3.43 & 3.42 \\
        \bottomrule
        \end{tabular}
    }
    \caption{\label{tab:appendix_data_sampling}
    ZeroDL performances according to the number of input data.
    }
    \end{table*}

\subsection{Additional Examples of Generated Classes}
    \begin{table*}[htbp]\centering\small
        \scalebox{0.85}{
        \begin{tabular}{c||c|c}
        \hlineB{3}
            \tt Data & \tt Method & \tt ClassLabels \\
        \hlineB{3}
        \multirow{2}{*}{\tt IMDB} & \tt ZeroDL &
            \begin{tabularx}{0.9\linewidth}{X}
                {{\bf Negative Sentiment}: {\color{gray}The list also includes various expressions of negative sentiment towards movies, films, shows, and documentaries. Some examples include (...)}} \\
                {{\bf Positive Sentiment}: \color{gray}{The list includes various expressions of positive sentiment towards movies, films, shows, and documentaries. Some examples include (...)}}
            \end{tabularx} \\
            \cline{2-3}
                & {\tt Gold} & \begin{tabularx}{0.90\linewidth}{X}
                {{\bf Negative}, {\bf Positive}}
            \end{tabularx} \\
            \cline{2-3}
        \hline
        \multirow{2}{*}{\tt SST-2} & \tt ZeroDL &
            \begin{tabularx}{0.90\linewidth}{X}
            {{\bf Positive Sentiment}: \color{gray}{All the sentiment labels that express a positive sentiment towards the movie, film, documentary, or subject. For example, (...)}} \\
            {{\bf Negative or Neutral Sentiment}: \color{gray}{All the sentiment labels that do not express a positive sentiment towards the movie, film, documentary, or subject. For example, (...)}}
            \end{tabularx} \\
            \cline{2-3}
                & {\tt Gold} & \begin{tabularx}{0.90\linewidth}{X}
                {{\bf Negative}, {\bf Positive}} \end{tabularx} \\
            \cline{2-3}
        \hline
        \multirow{3}{*}{\shortstack{\tt Yelp \\ \tt Rev}} & \tt ZeroDL & \begin{tabularx}{0.90\linewidth}{X}
        {{\bf Negative Sentiment}: \color{gray}{This class includes sentences expressing negative sentiments towards a place, food, or experience. Examples include (...)}} \\
        {{\bf Very Positive Sentiment}: \color{gray}{This class includes sentences expressing highly positive sentiments towards a place, food, or experience. Examples include (...)}} \\
        {{\bf Mixed Sentiment}: \color{gray}{This class includes sentences expressing mixed sentiments towards a place, food, or experience. Examples include (...)}} \\
        {{\bf Neutral Sentiment}: \color{gray}{This class includes sentences expressing neutral sentiments towards a place, food, or experience. Examples include (...)}} \\
        {{\bf Positive Sentiment}: \color{gray}{This class includes sentences expressing positive sentiments towards a place, food, or experience. Examples include (...)}}
        \end{tabularx} \\
            \cline{2-3}
                 & {\tt Gold} & \begin{tabularx}{0.90\linewidth}{X}
                 {{\bf Very Negative}, {\bf Negative}, {\bf Neutral}, {\bf Positive}, {\bf Very Positive}}\end{tabularx} \\
        \hline
        \multirow{3}{*}{\shortstack{\tt DBp \\ \tt (F)}} & \tt ZeroDL & \begin{tabularx}{0.90\linewidth}{X}
            {{\bf Aviation and Transportation}\color{gray}: This class includes topics related to aviation, aerospace technology, military history, maritime history, and transportation.} \\
            {{\bf Business and Economy}\color{gray}: This class includes topics related to business, finance, industries, companies, and economics.} \\
            {{\bf Sports and Biographies}\color{gray}: This class includes topics related to sports, athletes, and their biographies.} \\
            {{\bf Politics and Government}\color{gray}: This class includes topics related to politics, government, elections, and specific political parties.} \\
            {{\bf Education}\color{gray}: This class includes topics related to education, universities, schools, and specific educational institutions.} \\
            {{\bf History and Architecture}\color{gray}: This class includes topics related to history, architecture, historic sites, castles, and landmarks.} \\
            {{\bf Art and Entertainment}\color{gray}: This class includes topics related to art, music, entertainment, and specific artists or record labels.} \\
        \end{tabularx} \\
            \cline{2-3}
                 & {\tt Gold} & \begin{tabularx}{0.90\linewidth}{X}
                    {\bf Company}, {\bf EducationalInstitution}, {\bf Artist}, {\bf Athlete}, {\bf OfficeHolder}, {\bf MeanOfTransportation}, {\bf Building}
                 \end{tabularx} \\
        \hline
        \multirow{3}{*}{\shortstack{\tt DBp \\ \tt (B)}} & \tt ZeroDL & \begin{tabularx}{0.90\linewidth}{X}
        {{\bf Science and Technology}\color{gray}: This class includes topics related to paleontology, geology, volcanology, space exploration, and academic journals.} \\
        {{\bf Music and Entertainment}\color{gray}: This class includes topics related to music, album releases, jazz music, heavy metal music, hip hop music, and entertainment.} \\
        {{\bf Geography and Hydrology}\color{gray}: This class includes topics related to geography, hydrology, rivers, water bodies, and water resources.} \\
        {{\bf Botany and Plant Sciences}\color{gray}: This class includes topics related to botany, horticulture, plant taxonomy, plant conservation, and endangered species.} \\
        {{\bf Literature and Books}\color{gray}: This class includes topics related to literature, novels, fiction, mystery and crime fiction, and academic publications.} \\
        {{\bf Zoology and Entomology}\color{gray}: This class includes topics related to zoology, entomology, moths, butterflies, fish species, and arachnids.} \\
        {{\bf Film and Television}\color{gray}: This class includes topics related to film, cinema, movies, movie reviews, Bollywood, and television.}
        \end{tabularx} \\
            \cline{2-3}
                 & {\tt Gold} & \begin{tabularx}{0.90\linewidth}{X}
                    {{\bf NaturalPlace}, {\bf Village}, {\bf Animal}, {\bf Plant}, {\bf Album}, {\bf Film}, {\bf WrittenWork}}
                 \end{tabularx} \\
        \hline
        \multirow{3}{*}{\shortstack{\tt Yah \\ \tt (F)}} & \tt ZeroDL & \begin{tabularx}{0.90\linewidth}{X}
        {{\bf Personal Finance and Economics}\color{gray}: Topics related to personal finance, credit scores, debt management, taxes, and economics.} \\
        {{\bf Health and Wellness}\color{gray}: Topics related to health, medicine, fitness, nutrition, and wellness.} \\
        {{\bf Pop Culture and Entertainment}\color{gray}: Topics related to music, movies, TV shows, books, art, and entertainment.} \\
        {{\bf Technology and Computing}\color{gray}: Topics related to computers, technology, software, internet, telecommunications, and mobile phones.} \\
        {{\bf Miscellaneous}\color{gray}: Topics that do not fit neatly into any of the above categories, such as philosophy, religion, science, and humor.} \\
        {{\bf Fashion and Beauty}\color{gray}: Topics related to fashion, clothing, makeup, cosmetics, hair care, and beauty.} \\
        {{\bf Education and Careers}\color{gray}: Topics related to education, academic programs, scholarships, student loans, careers, and employment.}
        \end{tabularx} \\
            \cline{2-3}
                 & {\tt Gold} & \begin{tabularx}{0.90\linewidth}{X}
                    {{\bf ArtsAndHumanities}, {\bf BeautyAndStyle}, {\bf BusinessAndFinance},
{\bf ComputersAndInternet}, {\bf ConsumerElectronics}, {\bf EducationAndReference}, {\bf EntertainmnentAndMusic}}
                 \end{tabularx} \\
        \hline
        \multirow{3}{*}{\shortstack{\tt Yah \\ \tt (B)}} & \tt ZeroDL & \begin{tabularx}{0.90\linewidth}{X}
        {{\bf Food and Cooking}\color{gray}: This class includes topics related to various cuisines, recipes, food items, and cooking techniques.} \\
        {{\bf Home Improvement and DIY}\color{gray}: This class includes topics related to home repair, renovation, decorating, gardening, and DIY projects.} \\
        {{\bf Miscellaneous}\color{gray}: This class includes topics that do not fit neatly into any of the above categories, such as politics, education, art, and entertainment.} \\
        {{\bf Technology and Gaming}\color{gray}: This class includes topics related to video games, computer hardware, software, technology, and internet culture.} \\
        {{\bf Health and Wellness}\color{gray}: This class includes topics related to physical and mental health, nutrition, dieting, weight loss, fitness, exercise, and medical conditions.} \\
        {{\bf Animals and Pets}\color{gray}: This class includes topics related to various animals, pet care, animal rights, and wildlife.} \\
        {{\bf Religion and Philosophy}\color{gray}: This class includes topics related to various religions, theology, philosophy, and spirituality.}
        \end{tabularx} \\
            \cline{2-3}
                 & {\tt Gold} & \begin{tabularx}{0.90\linewidth}{X}
                    {{\bf FoodAndDrink}, {\bf GamesAndRecreation}, {\bf Health}, {\bf HomeAndGarden}, {\bf Pets}, {\bf PregnancyAndParenting}, {\bf SocietyAndCulture}}
                 \end{tabularx} \\
        \hlineB{3}
        \end{tabular}
    } 
    \caption{Additional examples of generated class labels. Class title is marked as bold and its description is colored as gray.}
    \label{tab:appendix_additional_class_gen}
    \end{table*}

    \begin{table*}[htbp] \centering\small
    \scalebox{0.9}{
        \begin{tabular}{c||c|c}
        \hlineB{3}
            {\tt Data} & {\tt TaskType} & \tt Generated Class Labels (Descriptions are omitted)\\
        \hlineB{3}
        \multirow{9}{*}{\tt IMDB} &  \tt sentiment & \begin{tabularx}{0.90\linewidth}{X}
            Negative, Very Positive, Highly Negative, Mixed, Neutral-Positive, Positive, Highly Positive, Neutral-Negative, Extremely Negative, Neutral; {\tt [Total 10]} \\
        \end{tabularx} \\
        \cline{2-3}
        & \tt topic & \begin{tabularx}{0.9\linewidth}{X}
            { Film and Television Influence on Business, Film and Television Influence on Politics, 
            Film and Television Production Companies, Film and Television Influence on Art, Film and Television Productions, Film Analysis or Review, Film and Television Influence on Society and Culture, Film and Television Influence on Technology, Film and Television Recommendations, Film and Television Awards, Film and Television, Film and Television Genres, Film and Television Influence on Society, Film and Television Marketing, Movie Reviews or Film Criticism, Film and Television Critics, Film and Television Technologies, Film and Television Festivals, Film or Media Criticism, Film and Television History, Film and Television Technology Trends, Film and Television Education, Film and Television Influence on Entertainment, Film and Television Influence on Education, Personal Opinion or Review, Film and Television Industry, Film and Television Distribution; {\tt [Total 27]}} \\
        \end{tabularx} \\
        \hline
        \multirow{4}{*}{\tt SST} &  \tt sentiment & \begin{tabularx}{0.90\linewidth}{X}
            The text expresses a negative sentiment towards the subject being described, The text expresses a negative or sad sentiment towards the film, The text expresses a negative or sad sentiment, The text has a negative tone, The text expresses a negative or cautionary sentiment, The text expresses a negative or slightly negative sentiment, Negative Sentiment, The text expresses a negative or slightly negative sentiment towards the film, The text expresses a negative sentiment, The text expresses a negative sentiment towards the movie being described, The text has a negative sentiment; {\tt [Total 11]} \\
        \end{tabularx} \\
        \cline{2-3}
        & \tt topic & \begin{tabularx}{0.9\linewidth}{X}
            Literature and Writing, Media and Entertainment, Mental Health and Psychology, Travel and Adventure, Film and Television, Family and Relationships, Science and Technology, Food and Cooking, Performing Arts, Sports, Education and Learning, Business and Finance, Greetings and Open-Ended Texts, Entertainment; {\tt [Total 14]} \\
        \end{tabularx} \\
        \hline
        \multirow{8}{*}{\tt \shortstack{\tt Yelp \\ \tt Rev}} &  \tt sentiment & \begin{tabularx}{0.90\linewidth}{X}
            Positive Sentiment, Highly Negative Sentiment, Neutral to Positive Sentiment, Negative Sentiment, Mixed Sentiment, Neutral Sentiment, Very Positive Sentiment, Neutral to Negative Sentiment; {\tt [Total: 8]} \\
        \end{tabularx} \\
        \cline{2-3}
        & \tt topic & \begin{tabularx}{0.9\linewidth}{X}
            Automotive, Environment or Nature, Legal or Law, Customer Service or Business, Nightlife or Entertainment, Education or Training, Sports or Fitness, Discrimination or Racism, Religion or Spirituality, Technology or Gadgets, Health or Medical, Travel or Tourism, Education or Learning, Beauty or Personal Care, Customer Reviews or Testimonials, Politics or Government, Food or Cooking, Food or Beverage, Shopping or Retail, Home Improvement or Construction, Science or Technology, Real Estate or Housing, Food Safety or Food Poisoning, Entertainment or Leisure, Arts or Culture, Personal Care or Beauty, Business or Economy, Personal Experiences, Dining Experience or Food Review; {\tt [Total 29]}
        \end{tabularx} \\
                \hline
        \multirow{2}{*}{\tt \shortstack{\tt AG \\ \tt News}} &  \tt sentiment & \begin{tabularx}{0.90\linewidth}{X}
            Negative: 12 expressions that contain negative sentiment, Positive: 25 expressions that contain positive sentiment, Negative: 12, Sentiment not clear: 3, Mixed: 11 expressions that contain a mix of positive and negative sentiment, Positive: 25, Neutral: 33, Neutral: 33 expressions that do not contain any clear positive or negative sentiment, Sentiment not clear: 3 expressions that do not provide enough context to determine a clear sentiment, Mixed: 11; {\tt [Total 10]} \\
        \end{tabularx} \\
        \cline{2-3}
        & \tt topic & \begin{tabularx}{0.9\linewidth}{X}
             \tt Same Result with Table~\ref{tab:class_info_analysis} \\
        \end{tabularx} \\
        \hline
        \multirow{4}{*}{\tt \shortstack{\tt DBp \\ \tt (F)}} &  \tt sentiment & \begin{tabularx}{0.90\linewidth}{X}
            Neutral, Positive, Negative; {\tt [Total 3]} \\
        \end{tabularx} \\
        \cline{2-3}
        & \tt topic & \begin{tabularx}{0.9\linewidth}{X}
            Baseball, Aviation, Sports and Biographies, History, Aircraft, Soccer or Football, People and Biographies, Ice Hockey, Education, American Football, General, Aircraft Design, Music or Entertainment, Higher Education or Universities, Healthcare or Hospitals, Football or Soccer; {\tt [Total 15]} \\
        \end{tabularx} \\
        \hline
        \multirow{6}{*}{\tt\shortstack{\tt DBp \\ \tt (B)}} &  \tt sentiment & \begin{tabularx}{0.90\linewidth}{X}
            Ambiguous, Neutral, Positive, Romantic, Negative, Mixed, Objective; {\tt [Total 6]} \\
        \end{tabularx} \\
        \cline{2-3}
        & \tt topic & \begin{tabularx}{0.9\linewidth}{X}
             Botany or Biology (specifically, Plant Science or Taxonomy), Botany or Endangered Species or Conservation Biology, Botany or Horticulture, Botany or Algae, Botany or Brazilian Flora,    Botany or Cacti, Botany or Mexican Flora, Botany or Tillandsia species, Botany or Orchids, Botany or Aquatic Plants, Botany or Plant Science, Botany or Tropical Plants, Botany or Hawaiian Flora, Botany or Plant Taxonomy, Botany or Palm Trees {\tt [Total 15]}
        \end{tabularx} \\
        \hline
        \multirow{4}{*}{\tt \shortstack{\tt Yah \\ \tt (F)}} &  \tt sentiment & \begin{tabularx}{0.90\linewidth}{X}
            Positive, Neutral, Mixed, Informational, Negative {\tt [Total 5]} \\
        \end{tabularx} \\
        \cline{2-3}
        & \tt topic & \begin{tabularx}{0.9\linewidth}{X}
            Miscellaneous (for topics that do not fit neatly into any specific category), Education and Careers, Makeup and Beauty, Gaming and Technology, Fashion and Clothing, Telecommunications, Housing and Real Estate, Music and Entertainment, Pop Culture and Entertainment, Personal Finance and Credit Scores; {\tt [Total 10]} \\
        \end{tabularx} \\
        \hline
        \multirow{6}{*}{\tt \shortstack{\tt Yah \\ \tt (B)}} &  \tt sentiment & \begin{tabularx}{0.90\linewidth}{X}
            Neutral; {\tt [Total 1]} \\
        \end{tabularx} \\
        \cline{2-3}
        & \tt topic & \begin{tabularx}{0.9\linewidth}{X}
            Health and Medical Concerns, Pop Culture and Entertainment, Mental Health and Psychology, Literature and Writing, Business and Finance, Food and Cooking, Video Games and Technology, Art and Creativity, Philosophy and Ethics, Education and Learning, Humor and Satire, Sports and Fitness, Religion and Theology, Home Improvement and DIY, Science and Technology, Travel and Adventure, Pets and Animals, Politics and Society, Pregnancy and Reproductive Health; {\tt [Total 19]} \\
        \end{tabularx} \\
        \hlineB{3}
        \end{tabular}
    }
    \caption{The examples of generated class labels when no constraints are given; we experiment with different task type and unlimited the number of clusters.}
    \label{tab:appendix_class_info_free}
    \end{table*}

\end{document}